%% file: main.tex
\pdfoutput=1
\newif\ifRAL

\RALtrue

\ifRAL

\documentclass[letter, 10pt, journal,twoside]{IEEEtran}      

\else

\documentclass[letterpaper, 10 pt, conference]{ieeeconf}  
\fi

\IEEEoverridecommandlockouts                              

\usepackage{wrapfig}
\usepackage{graphicx}
\usepackage{amsmath}
\usepackage{amssymb}
\usepackage{url}
\usepackage{hyperref}
\usepackage{subcaption}
\usepackage{color, colortbl}
\usepackage{algorithm}
\usepackage{algpseudocode} 
\usepackage[nolist]{acronym} 
\let\labelindent\relax
\usepackage[inline]{enumitem} 
\usepackage{cite}
\graphicspath{ {imgs/} }

\usepackage{pgfplots}
\pgfplotsset{compat=newest}
\usetikzlibrary{plotmarks}
\usetikzlibrary{arrows.meta}
\usepgfplotslibrary{patchplots}
\usepackage{grffile}

\usepackage{tikz}

\definecolor{Gray}{gray}{0.9}
\definecolor{Gray2}{rgb}{0.1,0.1,0.1}

\usepackage{draftwatermark}
    
\newif\ifshowComments
\showCommentstrue 

\ifshowComments
    \usepackage[draft,markup=bfit, deletedmarkup=sout, authormarkup=brackets]{changes}
\else
    \usepackage[final,markup=bfit, deletedmarkup=sout, authormarkup=brackets]{changes}
\fi

\definechangesauthor[name={Lukas},color=green]{LR}
\definechangesauthor[name={Matej}, color=orange]{MH}
\definechangesauthor[name={Jens}, color=cyan]{JL}
\definechangesauthor[name={Ville}, color=magenta]{VK}
\definechangesauthor[name={Jan}, color=teal]{JB}

\ifshowComments

\else

\fi

\input{acronyms.tex}

\input{macro.tex}

\SetWatermarkAngle{0}
\SetWatermarkColor{black}
\SetWatermarkLightness{0.5}
\SetWatermarkFontSize{10pt}
\SetWatermarkVerCenter{20pt}
\SetWatermarkText{\parbox{30cm}{%
  \centering This is the authors' final version of the manuscript published as\\
  \centering Rustler, L., Lundell, J., Behrens, J. K., Kyrki, V., \& Hoffmann, M. (2022). 'Active Visuo-Haptic Object Shape Completion'.\\
  \centering IEEE Robotics and Automation Letters 7 (2), 5254-5261. (C) IEEE. \url{https://doi.org/10.1109/LRA.2022.3152975} 
}}

\title{Active Visuo-Haptic Object Shape Completion}

\author{Lukas Rustler$^{1}$, Jens Lundell$^{2}$, Jan Kristof Behrens$^{3}$, Ville Kyrki$^{2}$, Matej Hoffmann$^{1}$ 
\ifRAL
 \thanks{Manuscript received: September, 9, 2021; Revised November, 19, 2021; Accepted January, 31, 2022.}%
 \thanks{This paper was recommended for publication by Editor Markus Vincze upon evaluation of the Associate Editor and Reviewers' comments. This work was supported by the project Interactive Perception-Action-Learning for Modelling Objects (IPALM) (H2020 --  FET -- ERA-NET Cofund -- CHIST-ERA III / Technology Agency of the Czech Republic, EPSILON, no. TH05020001 / Academy of Finland, no. 326304). M.H. and L.R. were additionally supported by OP VVV MEYS funded project CZ.02.1.01/0.0/0.0/16\_019/0000765 ``Research Center for Informatics''. L.R. was also supported by the Czech Technical University in Prague, grant no. SGS20/128/OHK3/2T/13. J.K.B. was supported by the European Regional Development Fund under project Robotics for Industry 4.0 (reg. no. CZ.02.1.01/0.0/0.0/15\_003/0000470).}
\fi
\thanks{$^{1}$ Lukas Rustler and Matej Hoffmann are with the Department of Cybernetics, Faculty of Electrical Engineering, CTU in Prague
 {\tt\small matej.hoffmann@fel.cvut.cz}}
\thanks{$^{2}$ Jens Lundell and Ville Kyrki are with the Intelligent Robotics Group, Department of Electrical Engineering and Automation, School of Electrical Engineering,
Aalto University, 02150 Espoo, Finland}
\thanks{$^{3}$Jan Kristof Behrens is with the Czech Institute of Informatics, Robotics, and Cybernetics, CTU in Prague}%
\thanks{Digital Object Identifier (DOI): see top of this page.}
}

\begin{document}

\maketitle

\begin{abstract}
Recent advancements in object shape completion have enabled impressive object reconstructions using only visual input. However, due to self-occlusion, the reconstructions have high uncertainty in the occluded object parts, which negatively impacts the performance of downstream robotic tasks such as grasping. In this work, we propose an active visuo-haptic shape completion method called Act-VH that actively computes where to touch the objects based on the reconstruction uncertainty. Act-VH reconstructs objects from point clouds and calculates the reconstruction uncertainty using IGR, a recent state-of-the-art implicit surface deep neural network. We experimentally evaluate the reconstruction accuracy of Act-VH against five baselines in simulation and in the real world. We also propose a new simulation environment for this purpose. The results show that Act-VH outperforms all baselines and that an uncertainty-driven haptic exploration policy leads to higher reconstruction accuracy than a random policy and a policy driven by Gaussian Process Implicit Surfaces. As a final experiment, we evaluate Act-VH and the best reconstruction baseline on grasping 10 novel objects. The results show that Act-VH reaches a significantly higher grasp success rate than the baseline on all objects. Together, this work opens up the door for using active visuo-haptic shape completion in more complex cluttered scenes.
\end{abstract}
\ifRAL
\fi
\input{Sections/introduction}
\input{Sections/related_work}
\input{Sections/method}

\input{Sections/experiments}

\input{Sections/conclusion}

\ifRAL
\else
\section{Acknowledgement}
This work was supported by the project Interactive Perception-Action-Learning for Modelling Objects (IPALM) (H2020 --  FET -- ERA-NET Cofund -- CHIST-ERA III / Technology Agency of the Czech Republic, EPSILON, no. TH05020001). M.H. and L.R. were additionally supported by OP VVV MEYS funded project CZ.02.1.01/0.0/0.0/16\_019/0000765 ``Research Center for Informatics''. L.R. was also supported by the Czech Technical University in Prague, grant no. SGS20/128/OHK3/2T/13. J.K.B. was supported by the European Regional Development Fund under project Robotics for Industry 4.0 (reg. no. CZ.02.1.01/0.0/0.0/15\_003/0000470).
\fi
\bibliographystyle{IEEEtran}
\bibliography{refs}

\end{document}

%% file: acronyms.tex
\newacro{dnn}[DNN]{Deep Neural Network}
\newacro{fcn}[FCN]{Fully Convolutional Network}
\newacro{pc}[PC]{Point Cloud}
\newacro{sdf}[SDF]{Signed Distance Function}
\newacro{cnn}[CNN]{Convolutional Neural Network}
\newacro{gnn}[GNN]{Graph Neural Network}
\newacro{dl}[DL]{Deep Learning}
\newacro{ml}[ML]{Machine Learning}
\newacro{gpis}[GPIS]{Gaussian Process Implicit Surface}
\newacro{mc}[MC]{Monte Carlo}
\newacro{mlp}[MLP]{Multi-Layer Perceptron}
\newacro{bpa}[BPA]{Ball Pivoting Algorithm}
\acrodefplural{gpis}[GPIS's]{Gaussian Process Implicit Surfaces}
\newacro{gpisp}[GPISP]{Gaussian Process Implicit Shape Potential}
\newacro{gp}[GP]{Gaussian Process}
\acrodefplural{gp}[GPs]{Gaussian Processes}

%% file: macro.tex
\newcommand{\equationref}[1]{\hyperref[#1]{Eq.~\ref*{#1}}}
\newcommand{\figref}[1]{\hyperref[#1]{Fig.~\ref*{#1}}}
\newcommand{\tabref}[1]{\hyperref[#1]{Table~\ref*{#1}}}
\newcommand{\secref}[1]{\hyperref[#1]{Section~\ref*{#1}}}
\newcommand{\algoref}[1]{\hyperref[#1]{Algorithm~\ref*{#1}}}

\newcommand{\norm}[1]{\left\lVert#1\right\rVert}
\newcommand{\Var}{\mathrm{Var}}

\DeclareMathOperator{\E}{\mathbb{E}}
\newcommand{\abs}[1]{\left\lvert#1\right\rvert}

\newcommand{\matr}[1]{\mathbf{#1}}
\newcommand{\argmax}{\operatornamewithlimits{argmax}}
\newcommand{\argmin}{\operatornamewithlimits{argmin}}
\newcommand*{\prob}{\mathsf{P}}

\def\methodname{Act-VH}

\def\ie{, \textit{i.e.}, }

\def\graspit{GraspIt!}
\def\pc{point cloud}
\def\pcs{point clouds}

%% file: Sections/introduction.tex
\section{Introduction}
\label{sec:introduction}

\ifRAL
\IEEEPARstart{S}{hape} 
\else
Shape 
\fi
completion, that is reconstructing the shape of an object based on incomplete sensory information, is an active research problem with many potential applications in medicine and robotics. To date, most methods have reconstructed objects from only visual data, including RGB images, depth images, or point clouds. The main drawback of visual data is that it is incomplete as the objects self-occlude\ie{}only the front side is visible from a single viewpoint. This increases the reconstruction uncertainty, which can negatively affect downstream robotic tasks such as grasping. A straightforward approach to combat the perceptual uncertainty is to gather additional data of the unseen object parts by touching the object and then reconstruct the object shape from combined visuo-haptic data. However, current visuo-haptic shape completion methods use heuristics to choose where to explore the objects \cite{varleyShapeCompletionEnabled2017} or require an impractical number of touches for a good reconstruction \cite{wang3DShapePerception2018a,bjorkmanEnhancingVisualPerception2013}. 

\begin{figure}[tb]
    \centering
    \includegraphics[width=0.49\textwidth]{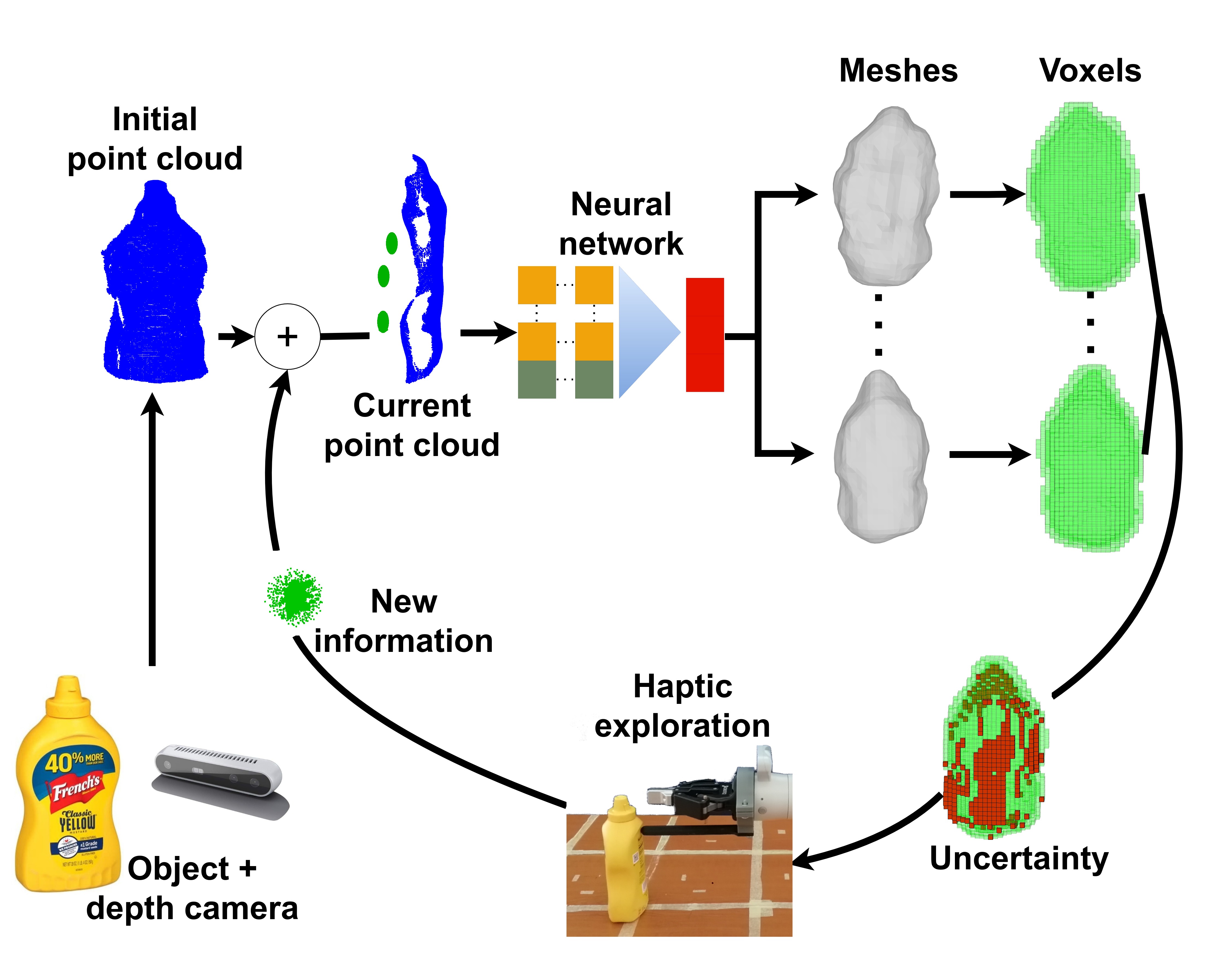}
    \caption{Schematic operation of \methodname{}. See text for details.}

    \label{fig:diagram}
    \vspace{-2em}
\end{figure}

We combat these issues in this work with \methodname{}, a data-efficient closed-loop active visuo-haptic shape completion method. The operation is schematically illustrated in \figref{fig:diagram}. First, a depth image is used to construct an initial point cloud. Then, a deep implicit surface network (IGR~\cite{groppImplicitGeometricRegularization2020}) generates several possible shape reconstructions by iteratively refining randomly initialized latent codes until the reconstructed shapes fit the input point cloud. The discrepancy between the reconstructed shapes are then used to form a single voxel-grid reconstruction with uncertainty. The voxel with the highest uncertainty is selected for haptic exploration, adding a new point to the object representation. This process is repeated, further refining the shape reconstruction.

We experimentally validated \methodname{} in simulation and the real world. To validate the method in simulation, we developed a new visuo-haptic benchmark task. Using this task, we compared the reconstruction accuracy of \methodname{} with random and uncertainty-driven exploration to five baselines on 105 reconstructions. The results showed that \methodname{} with uncertainty-driven exploration outperformed the baselines. In real-world experiments, we validated \methodname{} in terms of reconstruction accuracy and grasp success rates on 10 objects. Similar to the simulation results, the real-world reconstruction results also showed a significantly better accuracy using \methodname{}. Finally, the grasping experiment on the real robot showed that the grasp success rates after 5 touches increased from 30\% to 80\% using \methodname{}, while for the best reconstruction baseline it only increased from 20\% to 46.7\%, once again showing the benefits of \methodname{}.

The main contributions of this work are: 
\begin{enumerate*}[label=(\roman*)]
    \item a novel active visuo-haptic shape completion method;
    \item a visuo-haptic simulation environment; and
    \item an empirical evaluation of the proposed method against the state of the art, presenting improvements in terms of reconstruction accuracy (in simulation and on the real setup) and grasp success rates (real robot).
\end{enumerate*}
The simulation environment, which at the same time serves as a benchmark, and the data from experiments are available at {\footnotesize{\url{https://github.com/ctu-vras/visuo-haptic-shape-completion}}}. An accompanying video is here: {\footnotesize{\url{https://youtu.be/iZF4ph4zMEA}}}.

%% file: Sections/related_work.tex
\section{Related Work}
\label{sec:related_work}

An object shape can be reconstructed from visual input, haptic input, or their combination. 
Therefore, in this section, we split the review based on the sensory input used for reconstruction. Furthermore, the inputs can be collected only once or gathered actively to improve the reconstruction---the latter typically known as active perception \cite{bajcsy2018revisiting}. Thus, we also review active reconstruction methods per input modality.

\subsection{Visual-Only Shape Completion}

Completing object shapes from visual data is the most common approach because the data capture global information about the object. Early visual shape completion approaches were geometry- or template-based. Examples of geometry-based approaches reconstruct objects by mirroring them through their symmetry axis \cite{bohg2011mind} or using heuristics to fit primitives to resemble the object \cite{schnabel2009completion}. Template-based approaches search in a database for an object most similar to the perceived one \cite{pauly2005example}. The limitation of both methods is that they do not generalize well beyond specific objects. For instance, mirroring-based approaches result in poor reconstruction if the object has more than one axis of symmetry, while template matching will fail if the match is incorrect or no similar object exists in the database.  

To combat the limitations of geometry- and template-based methods, \ac{ml}-based shape completion approaches were proposed \cite{li2016dexterous,daiShapeCompletionUsing2017,hanHighResolutionShapeCompletion2017,varleyShapeCompletionEnabled2017,lundellRobustGraspPlanning2019,parkDeepSDFLearningContinuous2019,atzmonSALSignAgnostic2020,groppImplicitGeometricRegularization2020}. An early such approach trained a \ac{gpis} to reconstruct objects \cite{li2016dexterous}. However, the \ac{gpis} reconstructed overly smooth objects and, due to its poor scaling to many data points, the input point cloud had to be down-sampled, losing valuable information. More recent \ac{ml} approaches use \ac{dl} techniques to train 3D \acp{cnn} to complete the shape of objects represented as voxel grids \cite{daiShapeCompletionUsing2017,hanHighResolutionShapeCompletion2017,varleyShapeCompletionEnabled2017,lundellRobustGraspPlanning2019}. The limitation of voxel-based approaches is that the computation and memory requirements grow cubically with the object shape resolution. As such, fine object details are not preserved, which is essential when, for instance, sampling grasp proposals. 

To overcome the issue with voxel-grid representations, researchers proposed new network architectures that can handle continuous shapes \cite{parkDeepSDFLearningContinuous2019,atzmonSALSignAgnostic2020,groppImplicitGeometricRegularization2020}. The architectures  based on implicit surfaces are more computation- and memory-efficient than voxel-based representations and produce higher quality reconstructions. Because of these benefits, we chose to reconstruct objects with the implicit surface method IGR~\cite{groppImplicitGeometricRegularization2020}. 

Despite the impressive results of visual-only shape completion, the noise in the visual data and the objects' self-occlusions result in high reconstruction uncertainty, especially on the nonvisible parts of the object. If there is a possibility to move the camera, these limitations can be alleviated by actively choosing alternative viewpoints (also called next-best-view) \cite{wu20153d}. However, if the camera is not movable, another option is to use haptic data gathered by a robot.

\subsection{Haptic-Only Shape Completion}

If the robot has means to accurately detect and localize contacts with the object, tactile exploration can be more precise than visual data. Furthermore, any reachable part of the object---like its back side---can be explored. Most recent haptic-only shape completion approaches mainly reconstruct objects using classical \ac{ml} models such as implicit shape potentials \cite{ottenhausLocalImplicitSurface2016}, \acp{gp} \cite{yiActiveTactileObject2016}, \acp{gpis} \cite{driessActiveLearningQuery2017} or \acp{gpisp} \cite{dragiev2013uncertainty}. Additionally, some haptic exploration approaches actively explore the object to reduce the uncertainty in the reconstruction \cite{yiActiveTactileObject2016,driessActiveLearningQuery2017,dragiev2013uncertainty}. The limitation with haptic data is its local nature---one touch only explores a small object region. Consequently, accurate object reconstruction from tactile data requires tens \cite{dragiev2013uncertainty} to hundreds \cite{yiActiveTactileObject2016} of touches which is impractical for real robotic systems. 

\subsection{Visuo-Haptic Shape Completion}

To address the limitations of visual- or haptic-only shape completion, some works have proposed visuo-haptic shape completion  \cite{gandlerObjectShapeEstimation2020,ottenhausVisuoHapticGraspingUnknown2019, watkins-vallsMultiModalGeometricLearning2019,smith3DShapeReconstruction2020,smithActive3DShape2021,bjorkmanEnhancingVisualPerception2013,wang3DShapePerception2018a}. Most of these works reconstruct objects using \ac{ml} techniques such as \acp{gpis} \cite{gandlerObjectShapeEstimation2020,ottenhausVisuoHapticGraspingUnknown2019}, \acp{gp} \cite{bjorkmanEnhancingVisualPerception2013}, \acp{cnn} \cite{wang3DShapePerception2018a,watkins-vallsMultiModalGeometricLearning2019}, or \acp{gnn} \cite{smith3DShapeReconstruction2020,smithActive3DShape2021}. 
A limitation of the non-\ac{dl}-based visuo-haptic approaches \cite{gandlerObjectShapeEstimation2020,ottenhausVisuoHapticGraspingUnknown2019,bjorkmanEnhancingVisualPerception2013} is that good reconstructions often require haptic data all around the object. On the other hand, the main limitation of \ac{dl}-based \ac{cnn} approaches \cite{watkins-vallsMultiModalGeometricLearning2019,wang3DShapePerception2018a} is the low object resolution, while for \ac{gnn}s  \cite{smith3DShapeReconstruction2020,smithActive3DShape2021} it is the non-smooth shape reconstruction and that the reconstructions are only evaluated in simulation.

Another known problem for all visuo-haptic shape completion works is deciding where to explore the object haptically. One solution is to use heuristics, such as always approaching the object directly opposite the camera \cite{watkins-vallsMultiModalGeometricLearning2019}; another is to explore randomly \cite{smith3DShapeReconstruction2020}. Neither of these are particularly efficient as there exist more information-rich places to explore the object. To this end, some approaches learn where to explore the object \cite{smithActive3DShape2021} or use uncertainty of the reconstructions to guide  exploration \cite{bjorkmanEnhancingVisualPerception2013,wang3DShapePerception2018a,gandlerObjectShapeEstimation2020,ottenhausVisuoHapticGraspingUnknown2019}.

The work presented here also does uncertainty-driven visuo-haptic shape completion using \ac{dl}. Compared to similar works that use \acp{cnn} \cite{watkins-vallsMultiModalGeometricLearning2019,wang3DShapePerception2018a}, we use implicit surface networks to reconstruct highly detailed and smooth objects. Compared to works using \acp{gnn} \cite{smith3DShapeReconstruction2020,smithActive3DShape2021}, we evaluate our approach not only in simulation but also on real world reconstruction tasks. Furthermore, we propose a novel \ac{dl}-based uncertainty-driven exploration strategy and evaluate if our method benefits robotic grasping.

%% file: Sections/method.tex
\vspace{-1em}
\section{Method}
\label{sec:method}
We propose the method in \figref{fig:diagram} to do active visuo-haptic shape completion. We assume that the visual measurements are only captured once while the haptic measurements are collected incrementally by exploring the object. It is assumed that the object does not move after haptic exploration.
Based on these assumptions, the objective is to select a sequence of touches that would lead to the greatest improvements in reconstruction accuracy.
\vspace{-1em}
\subsection{Uncertainty-Driven Haptic Exploration}
\label{sec:prob_shape_comp}
Completing the shape of an object $O$ perfectly from real world measurements $Y$ is impossible due to the inherent noise and incompleteness of such measurements. The object $O$ can be modeled in several ways. In this work, we specifically use multiple of these representations as shown in \figref{fig:diagram}---from input point cloud, to \ac{sdf} as used by the IGR, to mesh, and finally voxel grid, where the uncertainty is computed.

We propose to model the object $O$ probabilistically as  
\begin{align}
    \label{eq:prob_object}
    \prob{}(O|Y),
\end{align}
where $O$ represents the occupancy of the object and $Y$ represents sensor measurements. The occupancy is represented as a voxel grid $O=(O^k)$ where $k$ is the index of a voxel such that $\prob{}(O^k)$ is the probability that voxel $k$ is part of the object. In this work, $Y$ consists of visual $v$ and haptic $h$ data. 

Formally, the objective is to, at each time step $t$, choose a location for haptic exploration that minimizes the uncertainty about the occupancy quantified as its variance
\begin{align}
    \label{eq:objective}
    \argmin_{h_t\in H} \Var(O_t|v,h_{1:t-1},h_t),
\end{align}
where $h_{1:t-1}$ is the data from previously executed haptic explorations and $H$ is the set of all possible haptic explorations. Note that the variance at time $0$ is based on visual data only $\Var(O_0|v)$.

Minimizing \equationref{eq:objective} requires a probabilistic model of the object's 3D shape, which is complex to form due to the high-dimensional nature of the data. Instead, we choose to approximate the model with a set of shape samples $o^{1:S}$ drawn from an underlying generative shape distribution $\prob{}(O_t|v,h_{1:t-1})$. The actual sampling process $o^{s} \sim \prob{}(O_t|v,h_{1:t-1})$ is described in the next section.

Assuming a set of shape samples are given in the form of voxel grids, we define the haptic exploration $h_t$ that minimizes \equationref{eq:objective} to be the voxel $k$ with the largest variance. This is formally expressed as
\begin{align}
    \label{eq:touch_location_var}
    \argmax_{k \in K} \Var(O^{k}),
\end{align}
where $k$ is a single voxel in a voxel grid $K$, and $\Var(O^k)$ is the variance of the shape samples $o^s$ for that voxel. Unfortunately, there often exist several voxels with the same variance and choosing one to explore is non-trivial. However, we found that most uncertain voxels form small clusters. We chose to explore the cluster with the most flat surface, which is advantageous for making a robust contact with the object.

\vspace{-1em}
\subsection{Sampling of Shapes}
\label{sec:sampling_of_shapes}
One of the crucial parts in \secref{sec:prob_shape_comp} is the sampling of shapes from the probability distribution $\prob{}(O_t|v,h_{1:t-1})$. Previous work on probabilistic shape completion \cite{lundellRobustGraspPlanning2019} achieved this by training a 3D \ac{cnn} to reconstruct voxelized objects and using the variational inference technique Monte Carlo dropout \cite{gal2016dropout} for sampling. However, for this process to work on visuo-haptic data, it requires training the \ac{cnn} on both haptic and visual data, with haptic data collected from random positions all around the object. Unfortunately, no such dataset exists, and curating one is expensive because of the haptic data collection process \cite{luo2017robotic}.  

Instead, we propose to train a reconstruction network that can accurately reconstruct shapes based on visuo-haptic data without explicitly training on such data. For this, we chose to use the IGR architecture \cite{groppImplicitGeometricRegularization2020} that learns the \ac{sdf} of the underlying surface. 
IGR is a \ac{mlp} $f(\mathbf{x};\boldsymbol{\theta},\cdot)$: $\mathbb{R}^3 \rightarrow \mathbb{R}$, where $\mathbf{x}$ is a 3D point and the parameters $\boldsymbol{\theta}$ are trained such that $f$ is approximately the \ac{sdf} to a plausible surface $\mathcal{M}$ defined by the point cloud $\mathcal{X}=\left\{\mathbf{x}_c\right\}_{c\in C}$ and optionally the point normals $\mathcal{N}=\left\{\mathbf{n}_c\right\}_{c\in C}$, where $C$ is the set of points in the \pc{}. The $\cdot$ is an additional parameter which is introduced below.
The loss function to train IGR is
\begin{align}
    \label{eq:rec_loss}
    \ell_{rec}(\boldsymbol{\theta},\cdot)=\ell_\mathcal{X}(\boldsymbol{\theta},\cdot)+\lambda \E_\mathbf{x}{\left[\norm{\nabla_\mathbf{x}f(\mathbf{x};\boldsymbol{\theta},\cdot)}-1\right]}^2,
\end{align}
where $\lambda>0$, and
\begin{align}
    \label{eq:X_loss}
\ell_\mathcal{X}(\boldsymbol{\theta},\cdot)=\frac{1}{\abs{C}}\sum_{c\in C} \left(\abs{f(\mathbf{x}_c;\boldsymbol{\theta},\cdot)} +\tau \norm{\nabla_\mathbf{x}f(\mathbf{x}_c;\boldsymbol{\theta},\cdot)-\mathbf{n}_c} \right).
\end{align}
The first term in \equationref{eq:rec_loss}, which is detailed in \equationref{eq:X_loss}, pushes f to vanish on $\mathcal{X}$. If normal data exist (our case), then $\tau:=1$ and $\nabla_{\mathbf{x}}f$ is pushed to the supplied normals $\mathcal{N}$. The second term in \equationref{eq:rec_loss}, called the Eikonal term, regularizes the network to produce smooth reconstructions by forcing the gradients of $\nabla_{\mathbf{x}}f$ to be of unit 2-norm.

By default, a separate IGR is trained for every single shape. However, this prohibits sampling from the underlying shape distribution $\prob{}(O_t|v,h_{1:t-1})$. Therefore, we chose to train a multi-shape IGR, which is realized by first selecting a separate latent vector $\mathbf{z}_j$ for each training example $j \in J$ and then train the network to approximate the \ac{sdf} associated with each $\mathbf{z}_j$. The multi-shape IGR takes the following form $f(\mathbf{x}; \boldsymbol{\theta},\mathbf{z}_j)$. 

To complete the shape of an object from a  partial point cloud with the multi-shape IGR comes down to finding a latent code $\hat{\mathbf{z}}_i$, where $i \in I$ are test samples, that best reconstructs the \ac{sdf} of the point cloud. To find such a latent code, we treat the observed point cloud as the ground truth and use gradient optimization to fine-tune an initially random code $\hat{\mathbf{z}}_{i,0}$. Formally, this fine-tuning is expressed as 
\begin{align}
    \label{eq:optimization_of_z}
    \hat{\mathbf{z}}_{i,t}=\hat{\mathbf{z}}_{i,t-1}-\alpha\nabla_{\hat{\mathbf{z}}_{i,t-1}} \ell(\boldsymbol{\theta},\hat{\mathbf{z}}_{i,t-1}),
\end{align}
where $\alpha$ is the step-size and $\nabla_{\hat{\mathbf{z}}_{i,t-1}}$ is the gradient of the following loss function: 
\begin{align}
    \ell(\boldsymbol{\theta},\hat{\mathbf{z}}_{i,t-1})= \ell_{rec}(\boldsymbol{\theta},\hat{\mathbf{z}}_{i,t-1})+\gamma\norm{\mathbf{z}_{i,t-1}}.
\end{align}
We chose $\gamma=0.01$, and $\ell_{rec}(\boldsymbol{\theta},\hat{\mathbf{z}}_{i,t-1})$ is the loss in \equationref{eq:rec_loss}. 

For drawing shape samples from a multi-shape IGR, two alternatives exist. The first, and most obvious, is to sample multiple latent codes $\hat{\mathbf{z}}_{1:S}$, where $S$ is the number of latent codes sampled, and optimize each of them individually for a fixed number of gradient descent steps. The second option, which we chose to use in our experiments, is to sample and optimize only one latent code $\hat{\mathbf{z}}_{1}$ and select $S$ intermediate optimized codes as the shape samples. For instance, if we optimized the latent code for 800 steps, we could select the latent code after 650, 700, 750 and 800 steps as our samples. This resembles Metropolis sampling in that samples are generated from a supposedly converged Markov chain, however, in our case we do not use the Metropolis rejection rule in the optimization process for simplicity but the stochasticity is introduced through sampling mini-batches in the optimization. This option is significantly faster than the first one and was empirically found to provide similar results.

\subsection{Active Visuo-Haptic Object Shape Completion}
\label{sec:uncertainty-driven}

We propose \algoref{alg:uncertainty-driven_shape completion} for active shape completion by combining the probabilistic shape completion from \secref{sec:prob_shape_comp} with the sampling of shapes in \secref{sec:sampling_of_shapes} (see also \figref{fig:diagram}). The algorithm starts by generating a random latent code $\hat{\mathbf{z}}_0$ (line \ref{op:init_latent}). Then that latent code is optimized with gradient descent over the current \pc{} and intermediate codes $\hat{\mathbf{z}}_g$ are saved (lines \ref{op:optimize_latent_code_begin}--\ref{op:init_new_iteration}). From the intermediate latent codes, meshes are reconstructed and transformed to voxel grids for the variance computation (line \ref{op:variance_computation}). Next, the voxel to touch (as described in \secref{sec:prob_shape_comp}) is computed, explored, and the information is added to the \pc{} (lines \ref{op:next_best_touch_location_begin}--\ref{op:next_best_touch_location_end}). Note that if no collision is detected at the target location (the robot is commanded to move on a straight line towards the target and 10 cm beyond), no \pc{} is saved and the robot returns to the start position and selects a new position for exploration. After all $M$ haptic explorations are done, the last latent code is optimized and the final shape is reconstructed (lines \ref{op:final_shape_reconstruction_begin}--\ref{op:final_shape_reconstruction_end}).
Some steps are illustrated in the accompanying video at {\footnotesize{\url{https://youtu.be/iZF4ph4zMEA}}}.

\subsection{Implementation details}
The IGR network was implemented in PyTorch 1.0.0. The network structure was the same as in \cite{groppImplicitGeometricRegularization2020} and consisted of 8 fully connected layers with 512 neurons and a skip connection in the 4th layer. The training was carried out on NVIDIA GeForce GTX 1080 Ti for 3500 iterations, with a batch size of 8 and a latent vector size of 256.

To train the network, we curated our own dataset of 87 unique meshes from the YCB~\cite{Calli2015} and Grasp Database~\cite{kappler2015leveraging} datasets. Each mesh was centered at the origin and scaled such that the longest dimension was between -1 and 1. To generate the ground truth point cloud of a mesh, we sampled 100000 points evenly over the complete object and, for each point, also estimated its normal. We rotated each mesh into 16 different views, resulting in 1392 training samples in total. Using the same procedure, we also generated a test set of 35 completely novel objects from both datasets.

\begin{figure}[tb]
\begin{algorithm}[H]
\caption{Active Visuo-Haptic Shape Completion}
\label{alg:uncertainty-driven_shape completion}
    \begin{algorithmic}[1] 
    \State \textbf{Inputs:} point cloud $\matr{P}$, number of haptic explorations M, number of gradient-descent steps G, steps before storing latent shape $L$
    \State \textbf{Output:} Final shape completion $O$ 
        \State $\matr{H} \gets \emptyset$ \Comment{Empty set of haptic data}
        \State $\matr{P}_0 \gets \matr{P}$ 
        \State $\hat{\mathbf{z}}_0 \gets \text{Sample initial latent code}$
        \label{op:init_latent}
        \For{$m\gets 1,~\dots,~M$}
            \State $\matr{Z} \gets \emptyset$ \Comment{Empty set of latent codes}
            \For{$g\gets 1,~\dots,~G$} \label{op:optimize_latent_code_begin}
                \State $\hat{\mathbf{z}}_g \gets \text{Optimize } \hat{\mathbf{z}}_{g-1} \text{ over } \matr{P}_{m-1} \text{ using \equationref{eq:optimization_of_z}}$ 
                \If {$g~\bmod{}~L == 0$}
                    \State $\matr{Z} \gets \matr{Z} + \hat{\mathbf{z}}_g$
                \EndIf
            \EndFor \label{op:optimize_latent_code_end}
            \State $\hat{\mathbf{z}}_0 \gets \hat{\mathbf{z}}_g$
            \label{op:init_new_iteration}
            \State $\matr{V} \gets \text{Reconstruct shapes from }\matr{Z} \text{ and calculate}\label{op:variance_computation} \newline ~~~~~~~~~~~~~ \text{their variance}$ \label{op:next_best_touch_location_begin}
            \State $h_m \gets \text{Calculate next touch using \equationref{eq:touch_location_var} and } \matr{V}$ \label{op:next_best_touch_location_begin}
            \State $\matr{H} \gets \text{Execute } h_m \text{ and append the touch point}$
            \State $\matr{P}_m \gets \matr{P}+\matr{H}$
            \label{op:next_best_touch_location_end}
        \EndFor
    \label{op:final_shape_reconstruction_begin}
    \For{$g\gets 1,~\dots,~G$}
        \State $\hat{\mathbf{z}}_g \gets \text{Optimize } \hat{\mathbf{z}}_{g-1} \text{ over } \matr{P}_I \text{ using \equationref{eq:optimization_of_z}}$ 
    \EndFor
    \State $O \gets \text{Reconstruct shape using }\hat{\mathbf{z}}_g$ \label{op:final_shape_reconstruction_end}
    \end{algorithmic}
\end{algorithm}
\vspace{-2.5em}
\end{figure}

%% file: Sections/experiments.tex
\section{Experiments}
\label{sec:experiments}

The experiments address the following two questions:
\begin{enumerate}
    \item What is the shape reconstruction accuracy of \methodname{}?
    \item What is the impact of \methodname{} on grasp success rate?
\end{enumerate}
To reliably answer these questions, we conducted two experiments. The first experiment (\secref{sec:exp_object_reconstruction}) evaluated shape reconstruction in simulation and in the real world, while the second experiment (\secref{sec:exp_robotic_grasping}) evaluated grasp success rates on real hardware.

\subsection{Experimental setup}

\begin{figure}[tb]
    \centering
    \vspace{1em}
    \begin{subfigure}[t]{0.48\textwidth}\caption{}\end{subfigure}
    \addtocounter{subfigure}{1}%
    \begin{subfigure}[t]{0.48\textwidth}
        \centering
        \vspace{-2em}%
        \includegraphics[height=0.3\textheight]{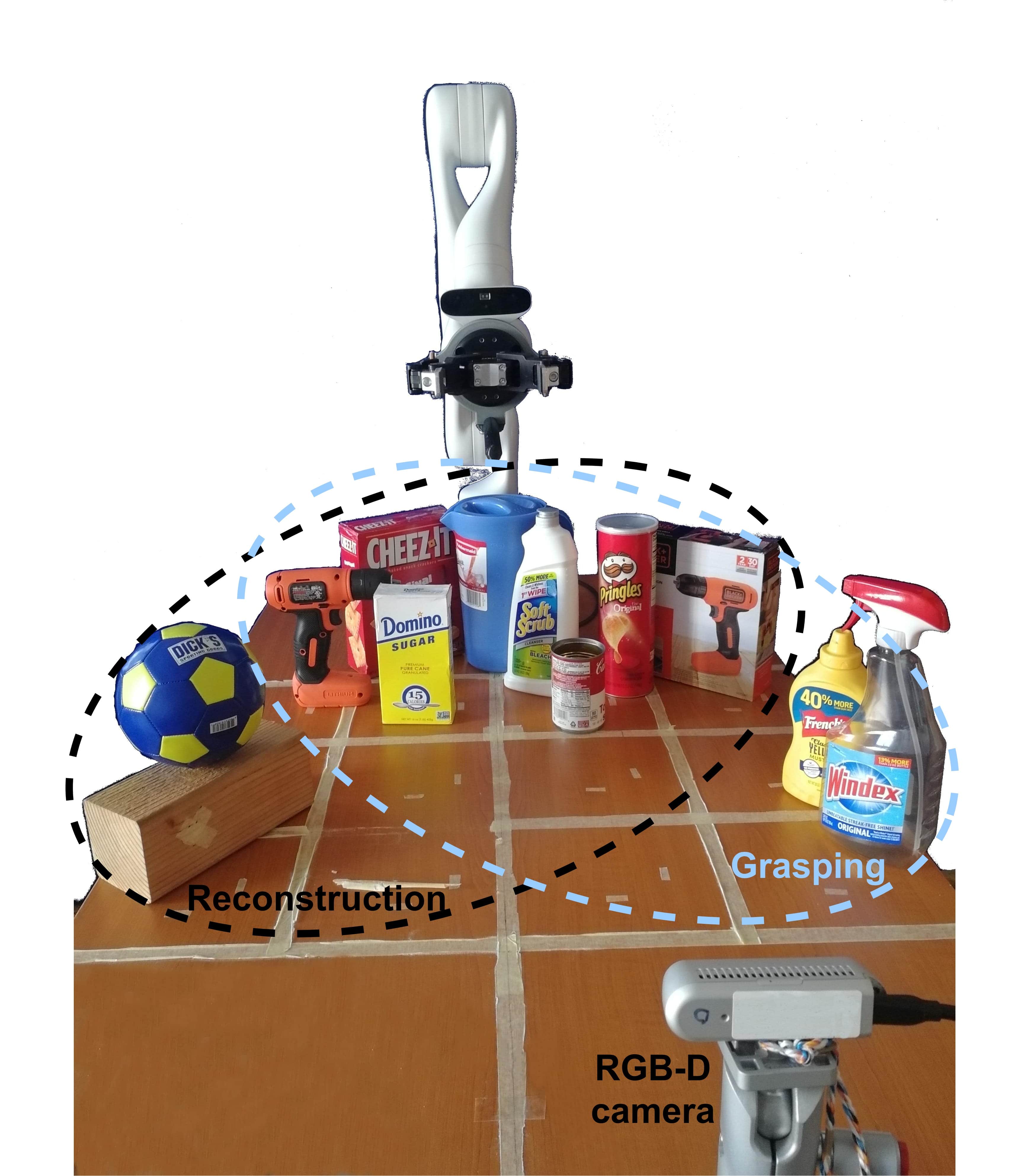}
        \caption{The real-world setup with the Kinova Gen3 robotic arm, the RGB-D camera (always in front of the robot), and the objects used for reconstruction and for grasping.}
        \label{fig:real_setup}
    \end{subfigure}
    \addtocounter{subfigure}{-3}
    \begin{subfigure}[t]{0.185\textwidth}
        \centering
        \vspace{-23.5em}%
        \includegraphics[width=1\textwidth]{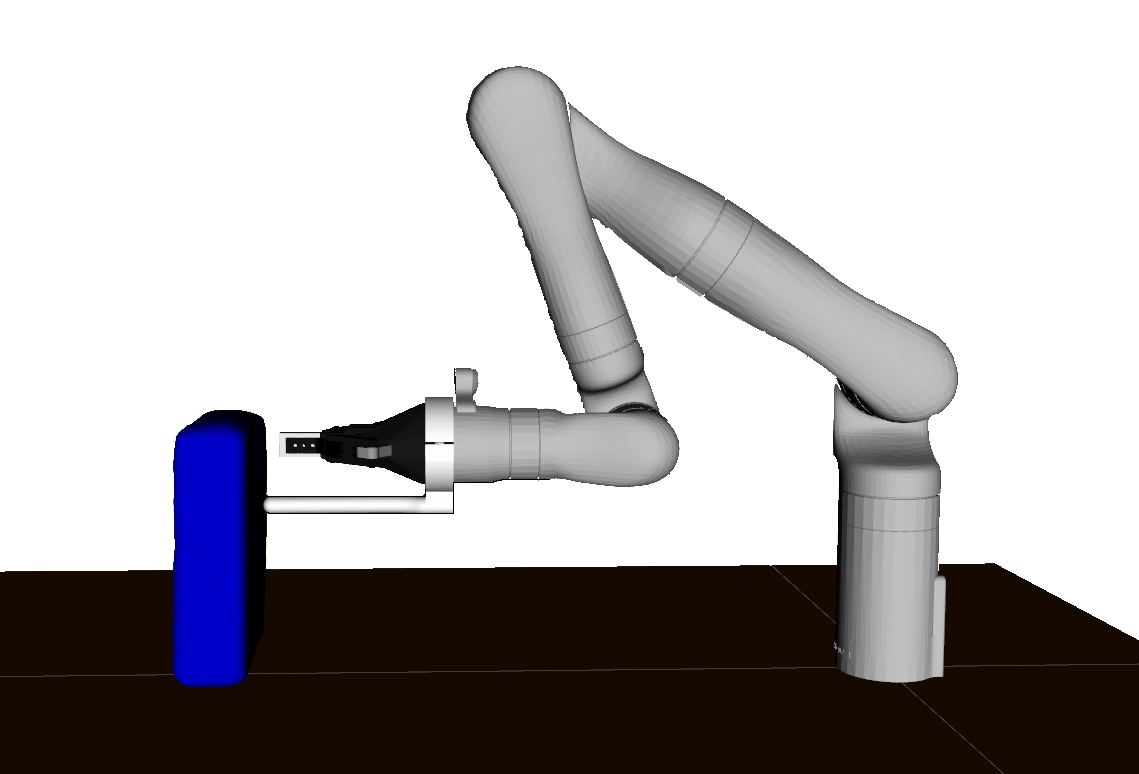}
        \caption{Simulation.}
        \label{subfig:simulation}
    \end{subfigure}\hspace{5.5em}%
    \begin{subfigure}[t]{0.18\textwidth}
        \centering
        \vspace{-24em}%
        \includegraphics[width=1\textwidth]{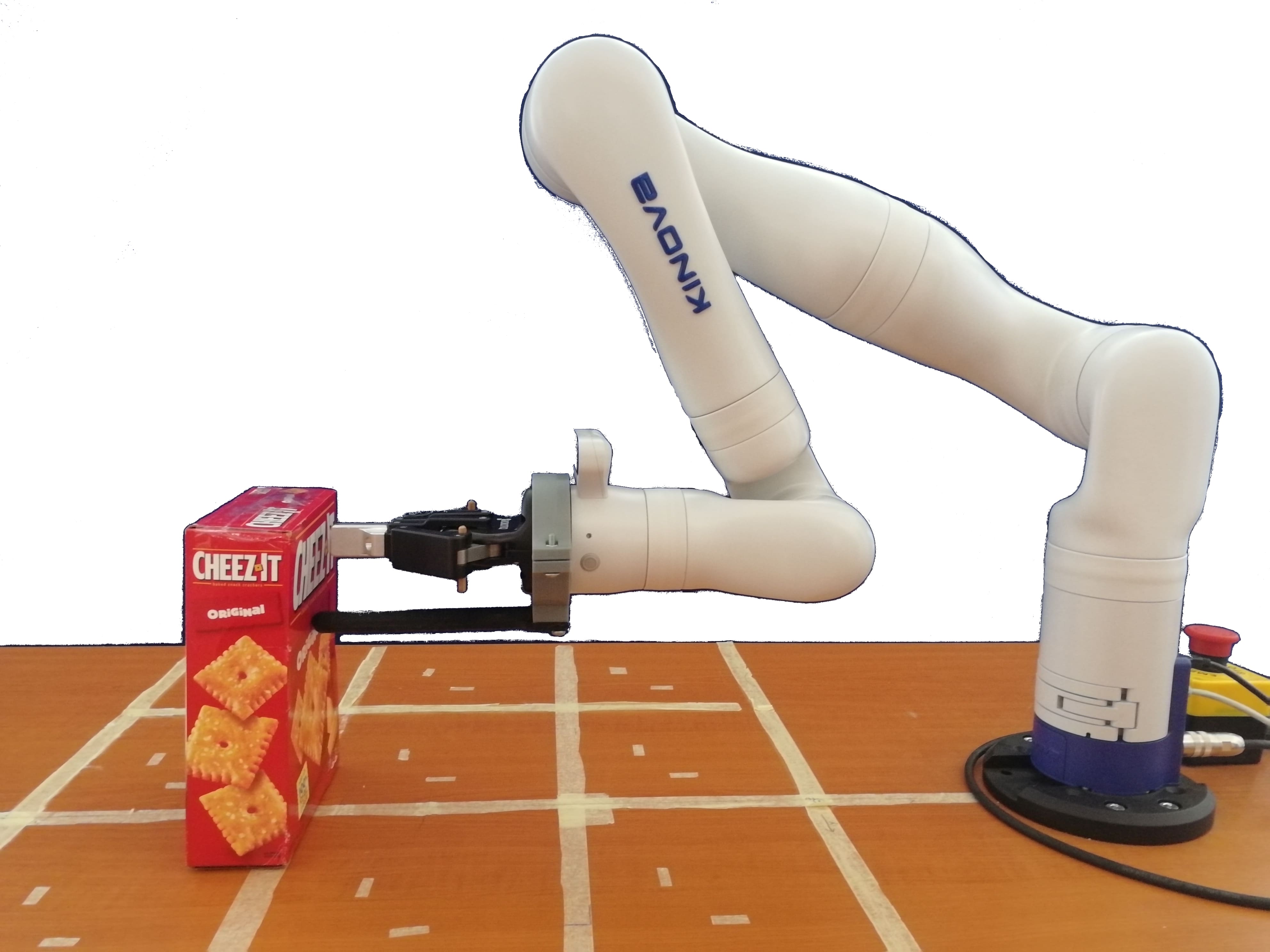}
        \caption{Real world.}
        \label{subfig:real}
    \end{subfigure}
    \caption{Simulated (a) and real (b,c) environment.}
    \label{fig:environments}
    \vspace{-1.5em}
\end{figure}

In both the simulation and real-world experiments, we used a Kinova Gen3 robot equipped with a custom made finger to perform haptic exploration and a Robotiq 2F-85 gripper for grasping. In the real-world experiments, we used an Intel RealSense D430 depth camera to capture the point cloud -- see \figref{fig:environments}. 

For haptic exploration, we moved the robot along a pre-defined approach direction until the torques of the robot joints crossed a pre-defined threshold. The global position of the finger was then transformed into the same reference system as the visual point cloud. The objects were attached to the table using a double-sided tape. If nothing specific is noted, we maximally executed five touches in all experiments to keep the total execution time low. The time taken to run the pipeline with five touches is about 5 minutes on average.

In the reconstruction experiments, we benchmarked \methodname{} against five other baseline methods: \ac{bpa} \cite{bernardiniBallpivotingAlgorithmSurface1999}, Poisson reconstruction (Poisson) \cite{kazhdan2006poisson}, Convex Hull reconstruction (Hull), Alpha shapes (Alpha) \cite{guoSurfaceReconstructionUsing1997}, and \ac{gpis} \cite{Williams2006}. \ac{bpa} \cite{bernardiniBallpivotingAlgorithmSurface1999} reconstructs a shape by rolling a sphere with a pre-defined radius over all points, and if three points are inside the sphere, they are connected with a triangle. Poisson reconstruction \cite{kazhdan2006poisson} solves an optimization problem that creates a smooth surface over the points but can only reconstruct the visible part of the surface. The Hull method reconstructs the input point cloud with a convex hull over the points. Alpha is a generalized Hull method that smooths the object surface and can also remove volume from the inside of concave objects. \ac{gpis} \cite{Williams2006} trains a \ac{gp} to reconstruct the implicit surface of the object. For \ac{gpis}, we used similar hyper-parameters as reported in \cite{watkins-vallsMultiModalGeometricLearning2019}.

To benchmark other methods, we still used \methodname{} to calculate the reconstruction uncertainty and where to touch the object but used a baseline method instead of IGR for the final reconstruction. In the reconstruction experiments, we evaluated all methods using the Chamfer distance and Jaccard similarity, while in the grasping experiment, we used the grasp success rate. For calculating the Jaccard similarity, each mesh was voxelized into a voxel grid of size $40^3$.

\subsection{Object Reconstruction}
\label{sec:exp_object_reconstruction}

In this experiment, we evaluated the reconstruction accuracy in simulation and in the real world. To evaluate in simulation, we developed our own visuo-haptic robotic simulation environment in the MuJoCo physics simulator \cite{Todorov2012}.
The environment, shown in \figref{subfig:simulation}, consist of a robot, an object mesh, and a virtual camera for capturing the point cloud of the object. In both simulation and the real word, the robot planned and moved to the location we wanted to haptically explore and stopped once contact was detected as shown in \figref{subfig:real}. 

In simulation, we evaluated the reconstruction accuracy on 35 test objects, while in the real world, we used the 10 objects shown in \figref{fig:real_setup} that were selected because they differed in size and shape. Each reconstruction was repeated three times for each object and method combination, resulting in 105 unique reconstructions per method in simulation and 30 in the real world. In the simulation experiment, we further compared \methodname{} to: (i) a random policy that touched the first reachable voxel from the set of uniformly sampled voxels on the surface of the reconstruction, and (ii) a \ac{gpis}-driven policy, where the voxel with the largest standard deviation was selected~\cite{yiActiveTactileObject2016}.

Note that \ac{gpis} approximates the surface covered by the input points but does not assume a closed volume. Therefore, we needed to select a reasonable first touch point heuristically.

\begin{figure}[tb]
    \centering
    \includegraphics[width=0.44\textwidth]{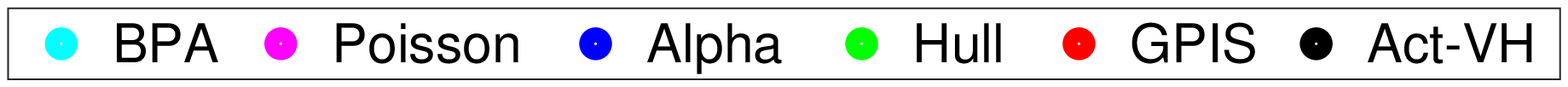}
    \begin{subfigure}[t]{0.48\textwidth}
        \centering
        \includegraphics[width=0.49\textwidth,trim={0 0 1.0cm 0},clip]{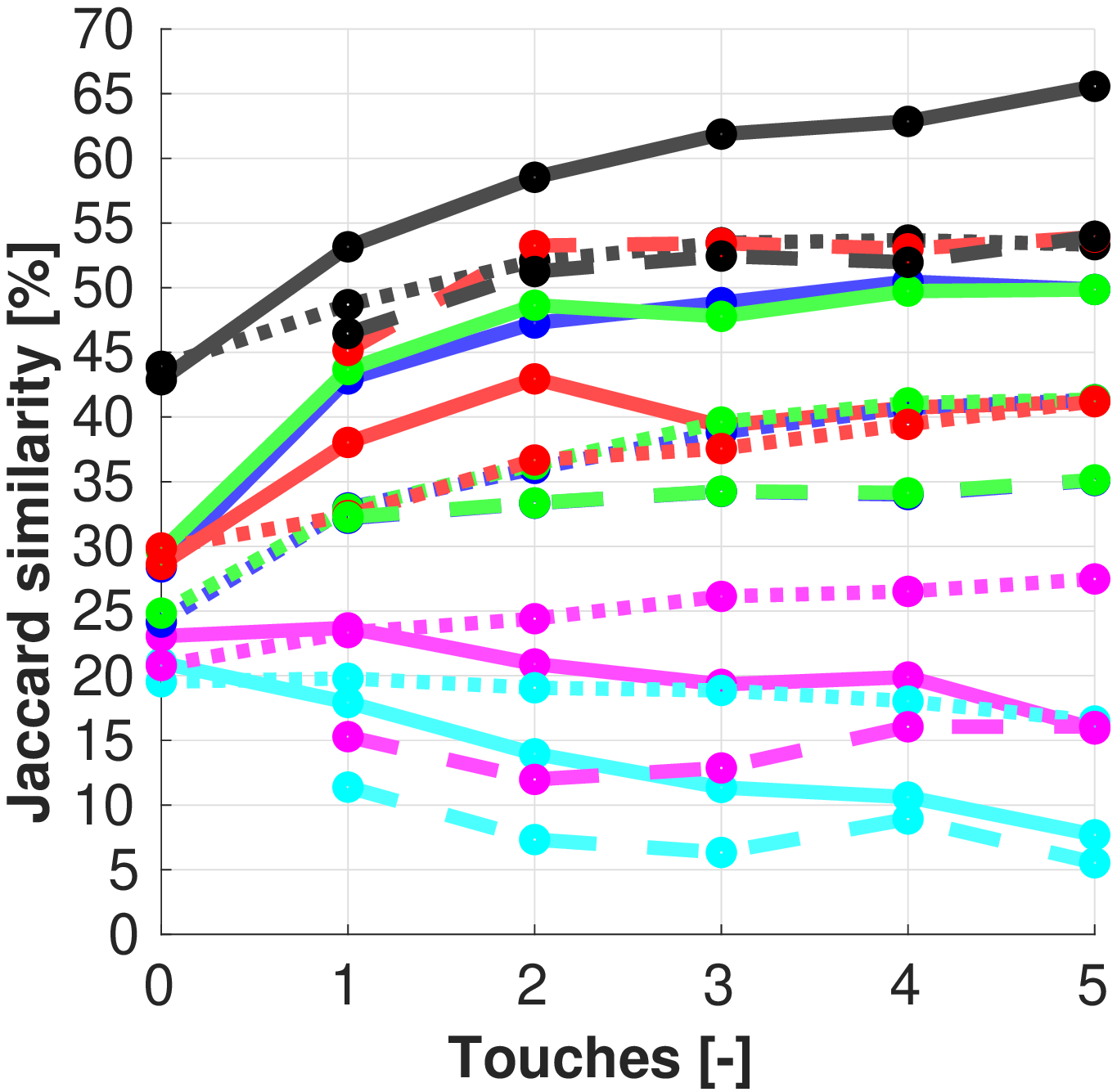}
        \includegraphics[width=0.49\textwidth,trim={0 0 1.0cm 0},clip]{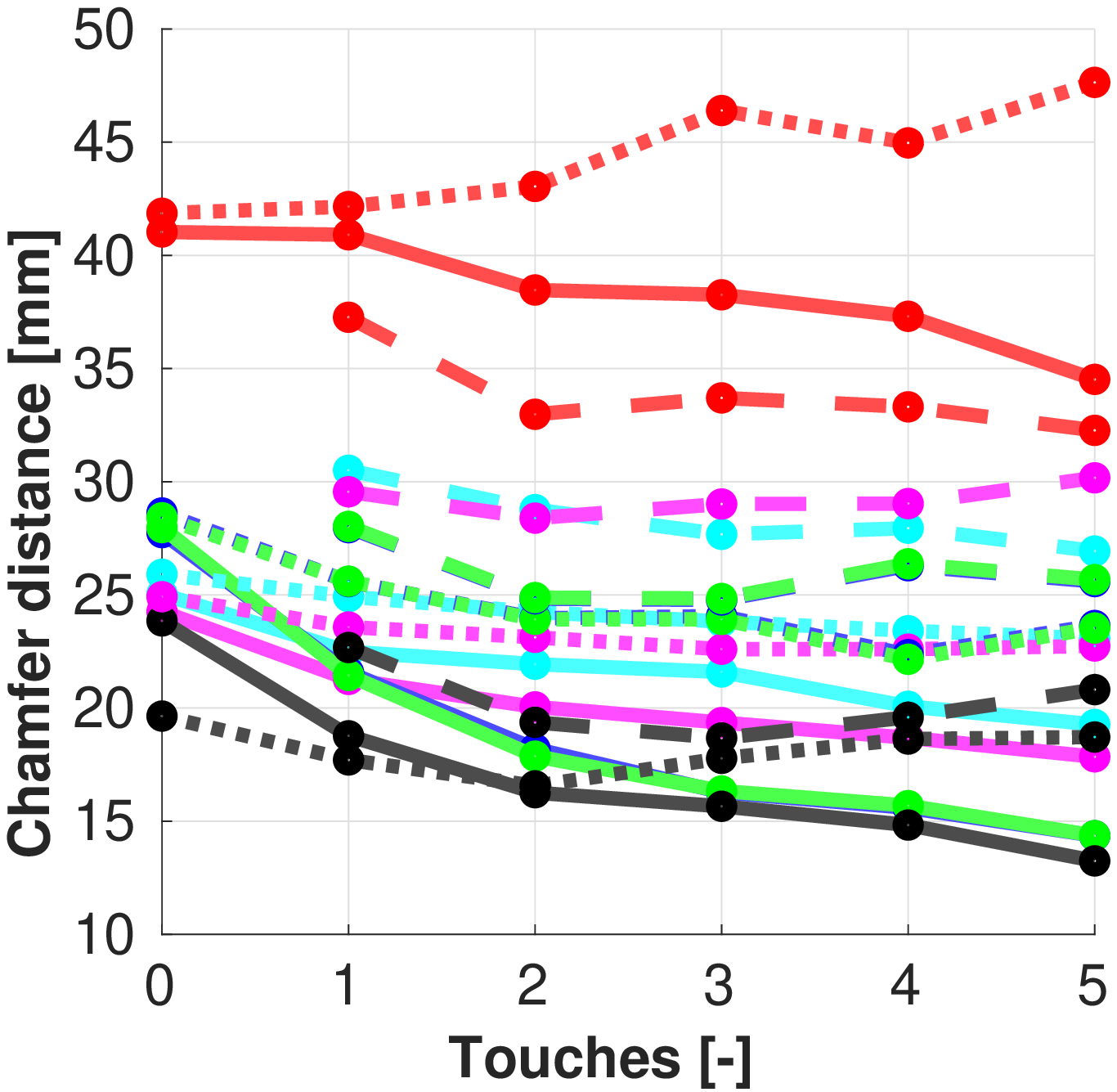}
        \caption{Simulation results. Solid lines represent the accuracy obtained when executing \methodname{} touches, dotted lines random touches, and dashed lines  
        \ac{gpis}-uncertainty-driven touches. 
       \label{subfigure:reconstruction_res_sim}}
    \end{subfigure}
    \begin{subfigure}[t]{0.48\textwidth}
        \centering
        \includegraphics[width=0.49\textwidth,trim={0 0 1.0cm 0},clip]{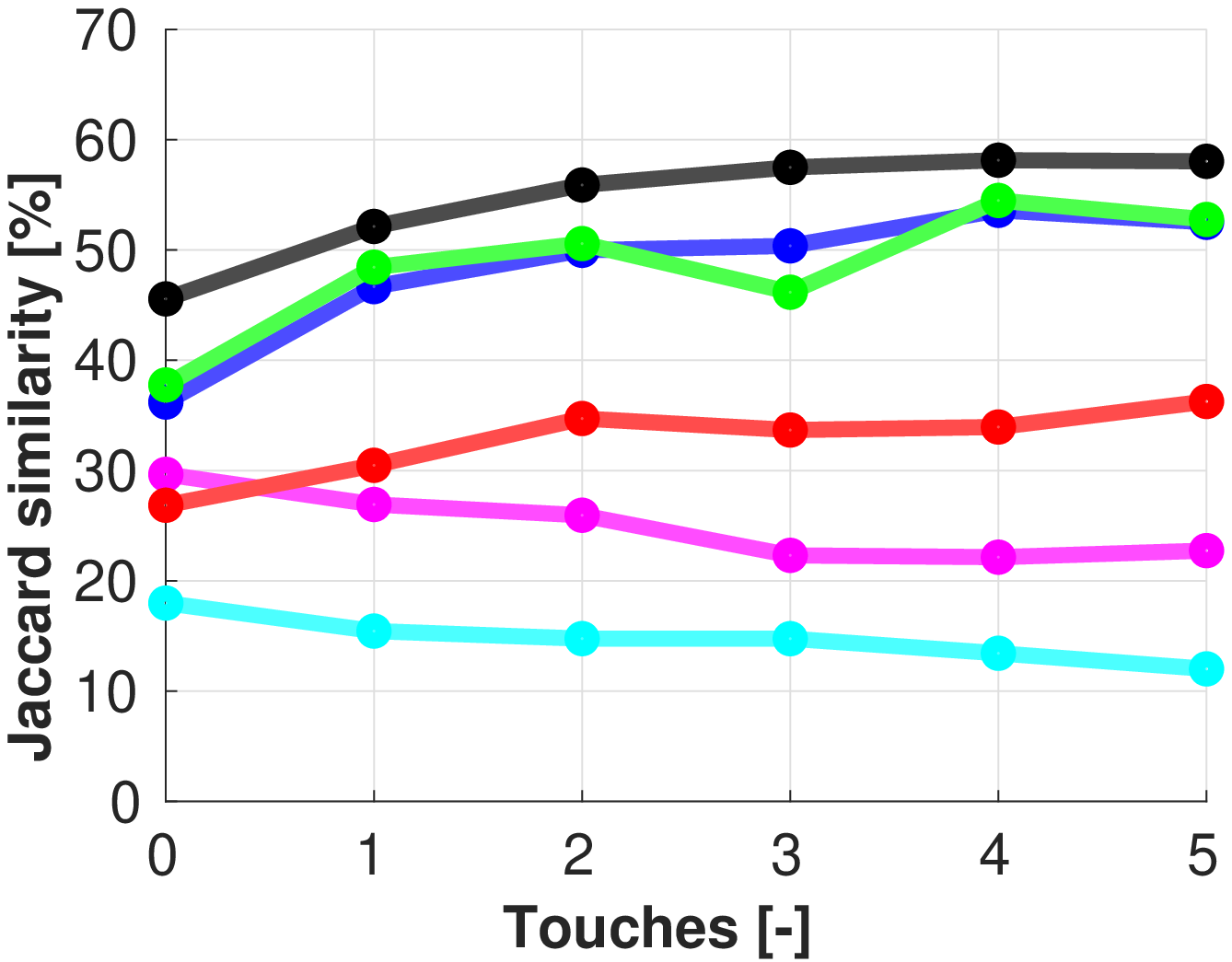}
        \includegraphics[width=0.49\textwidth,trim={0 0 1.0cm 0},clip]{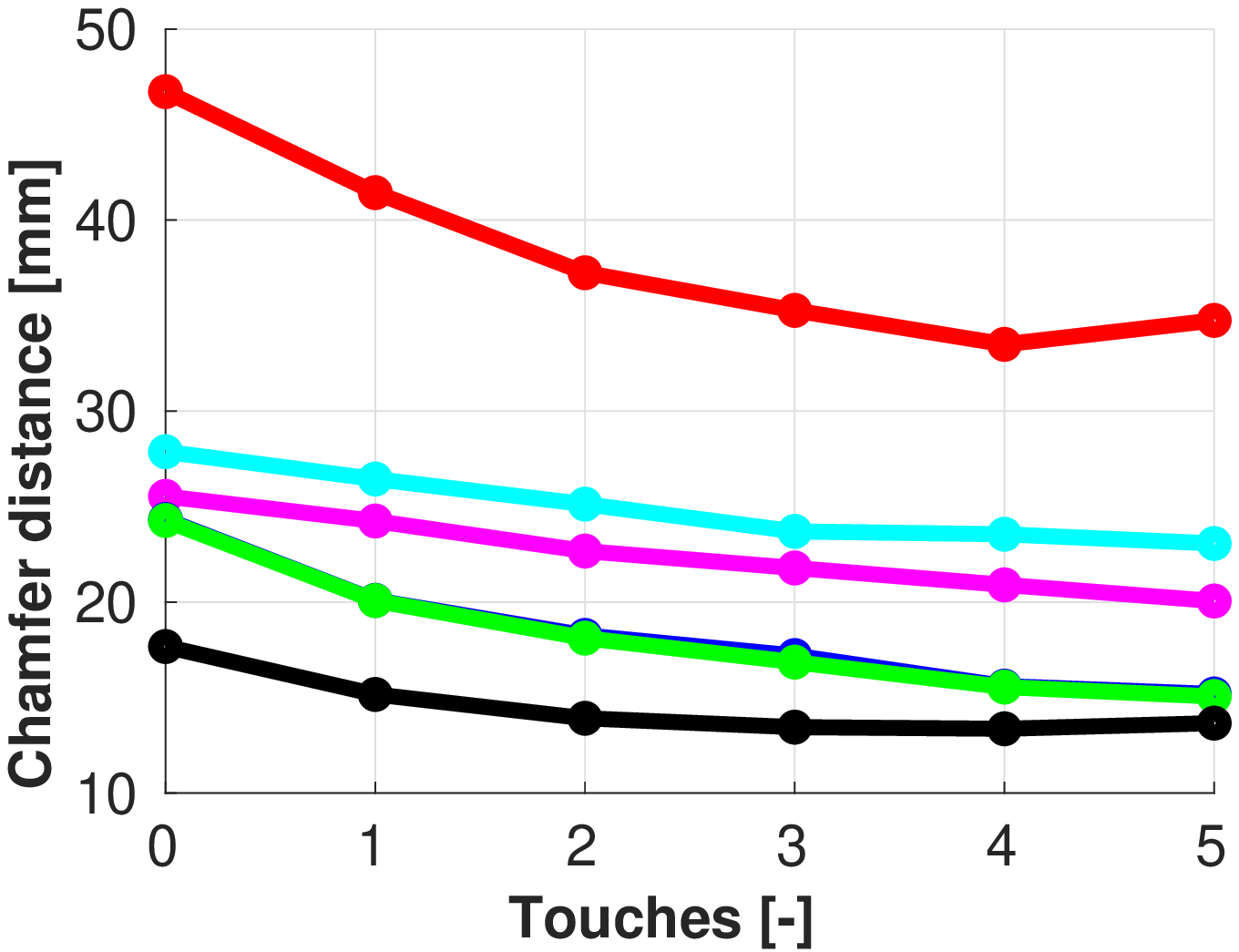}
        \caption{Real world results.\label{subfigure:reconstruction_res_real}}
    \end{subfigure}
    \caption{The average reconstruction accuracy from simulation (a) and the real world (b). For Jaccard similarity, larger values are better; for Chamfer distance, smaller values are better.}
    \label{fig:comparison_sep_sim_and_real}
    \vspace{-2em}
\end{figure}

\figref{fig:comparison_sep_sim_and_real} shows the reconstruction results separately for simulation (\figref{subfigure:reconstruction_res_sim}) and real world (\figref{subfigure:reconstruction_res_real}). Overall, the reconstruction accuracy for all methods improved with the number of touches, meaning that Jaccard similarity increased and Chamfer distance decreased. Furthermore, both simulation and real world results follow the same trend indicating the robustness of our method. 

Based on the results in \figref{fig:comparison_sep_sim_and_real}, we can clearly see that \methodname{} outperforms all other baselines across the board. For example, \figref{subfigure:reconstruction_res_sim} shows that \methodname{} outperforms the random one already after one touch, and after five touches \methodname{} reaches around 10-20\% higher Jaccard similarity than the random one and over 5 mm lower Chamfer distance. 

The results in \figref{subfigure:reconstruction_res_sim} also show that \methodname{} exploration outperforms \ac{gpis} exploration. The reason \ac{gpis} performs poorly is because it requires touches to be evenly distributed around the whole object.\footnote{Note that we used the exponential kernel. Results with other kernels (\textit{e.g.}, thin plate kernel as in \cite{bjorkmanEnhancingVisualPerception2013}) may be different.} 
Other reconstruction works presented similar results \cite{yiActiveTactileObject2016,watkins-vallsMultiModalGeometricLearning2019}, where in the haptic-only work \cite{yiActiveTactileObject2016}, more than 100 touches were required to reconstruct the front side only, while in the visuo-haptic work \cite{watkins-vallsMultiModalGeometricLearning2019}, more than 20 touches were collected and the results were still poor.

When comparing reconstruction methods, the second best method was Hull, with a Chamfer distance around 15 mm. One reason Hull achieved such a low Chamfer distance is that it creates more sharp reconstructions, which strongly influences the Chamfer distance. However, for Hull to produce good reconstructions the \pc{} must contain points covering the whole object, which is in practice best achieved with \methodname{} exploration.

Interestingly, with \methodname{} exploration, the Jaccard similarity of \ac{bpa} and Poisson decreased with more touches, while with a random policy, it stayed flat or increased slightly. These results point to the fact that \ac{bpa} and Poisson can only reconstruct the visible part of the surface\ie{}parts that are covered by the point cloud, and random exploration has a high chance of being close to those points resulting in more useful information. In contrast, \methodname{} exploration most likely returns a data point far from the visual point cloud, leading \ac{bpa} and Poisson to create strange artifacts that, once voxelized, result in lower Jaccard similarity. Although the \methodname{} policy resulted in lower Jaccard similarity for \ac{bpa} and Poisson, it still outperformed the random policy in terms of Chamfer distance.

\begin{figure*}[tb]
    \centering
    \includegraphics[width=0.7\textwidth]{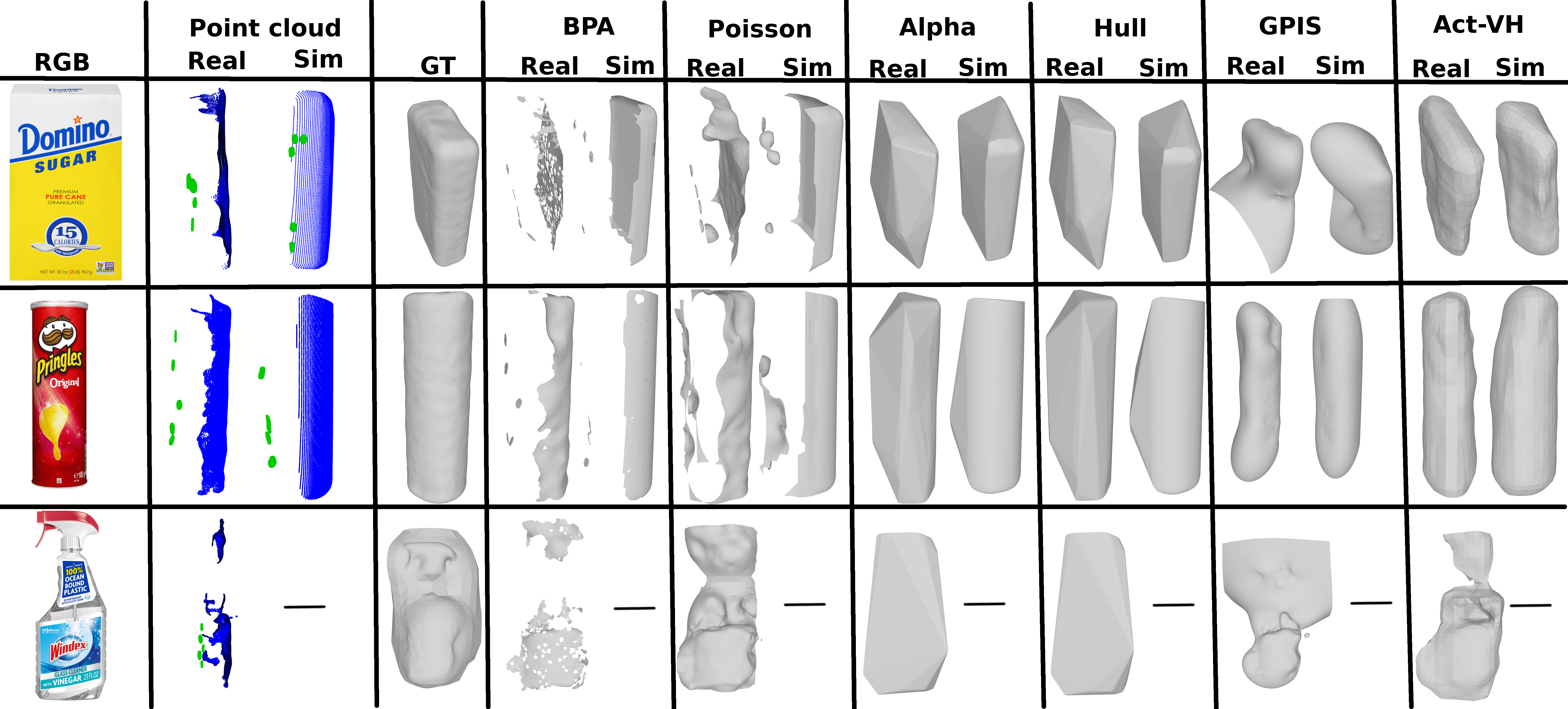}
    \caption{Reconstruction examples in simulation (Sim) and real with all methods on three objects: A basic rectangular object (upper row), the object for which \methodname{} achieved the worst grasp success rate  (middle row), and an adversarial object (bottom row). Touches in the \pcs{} are highlighted with green color. There is no reconstruction in simulation for the adversarial object because there was no appropriate ground truth mesh.}
    \label{fig:meshes}
    \vspace{-1em}
\end{figure*}

\begin{figure*}[tb]
    \centering
    \includegraphics[width=0.7\textwidth]{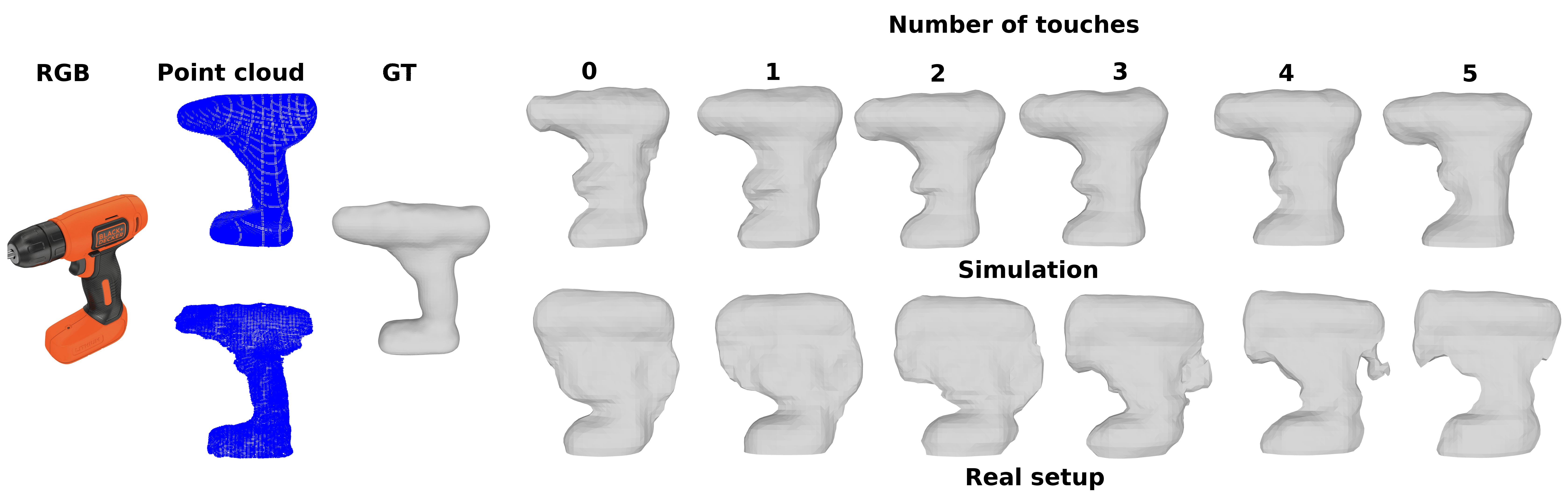}
    \caption{An example reconstruction of an object after each touch with \methodname{}  in simulation (upper row) and the real world (bottom row).}
    \label{fig:meshes_touches_both}
    \vspace{-2em}
\end{figure*}

\figref{fig:meshes} shows example reconstructions after 5 touches. These examples show that: (i) \ac{bpa} and Poisson are unable to complete the whole object accurately, (ii) Alpha and Hull reconstruct very sharp and unrealistic objects, and (iii) \ac{gpis} is poor at reconstructing the object where no points are available. In contrast, \methodname{} can capture both global and local features, resulting in smooth and faithful reconstructions. A challenging object to shape complete was the partly transparent spray bottle in the bottom row of \figref{fig:meshes}. Nevertheless, \methodname{} still reconstructed it quite well compared to the ground truth and other methods. The results of an incremental \methodname{} reconstruction with five touches is visualized in \figref{fig:meshes_touches_both}, highlighting that if the first reconstruction is good, which happened to be the case in simulation, additional touches only locally refine the objects. However, if the initial reconstruction is poor, which was the case in the real world experiment, additional touches lead to more global refinements. Note that the initial estimation could be improved by replacing random sampling for mini-batches with Farthest Point Sampling (FPS) (as in \cite{qi2017PointNet}) which better preserves the global information about the object. However, our experiments showed that after haptic exploration, FPS is not leveraging this information well and is outperformed by random selection.

\begin{wrapfigure}[14]{I}{0.24\textwidth}
    \vspace{-2em}
    \begin{center}
        \includegraphics[width=0.24\textwidth]{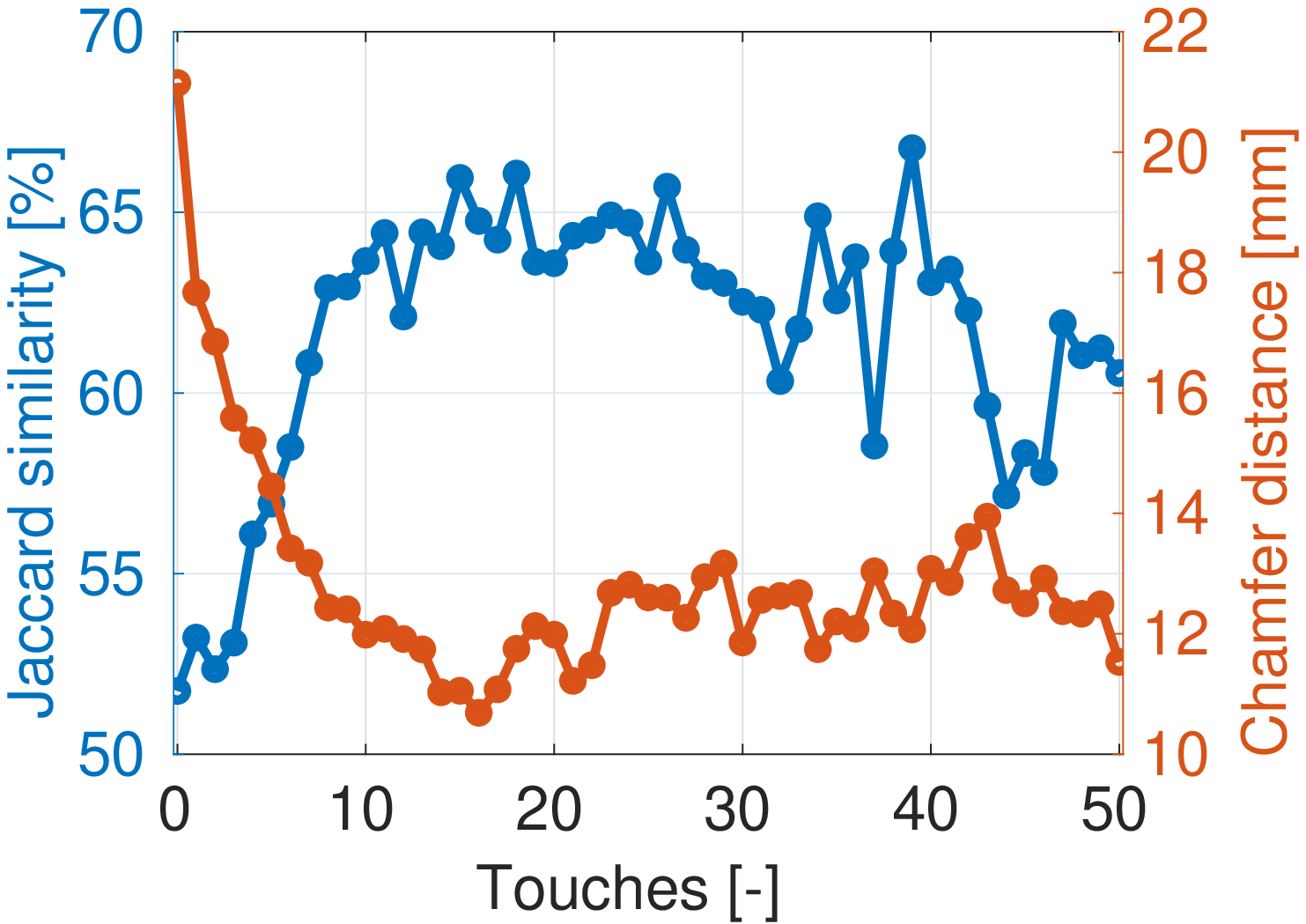}
    \end{center}
    \caption{Long exploration -- 50 touches in simulation. Average Jaccard similarity and Chamfer distance for three objects over three repetitions.} 
    \label{fig:fifty}
\end{wrapfigure}

Finally, we investigated if the reconstruction accuracies in \figref{fig:comparison_sep_sim_and_real} approach some steady-state value with more touches. We let \methodname{} explore three objects three times in simulation with 50 touches.
The results are presented in \figref{fig:fifty}. Based on these results, it seems that \methodname{} does approach a steady-state Chamfer distance after about 20 touches, albeit some fluctuations are still present. The same conclusion cannot be made for the Jaccard similarity, which actually gets worse after about 25 touches. The primary reason the Jaccard similarity starts to decrease was due to errors in the exact location of contact which originated from imprecise joint torque collision detection. Although \methodname{} can cope with some errors, ultimately, after enough touches, the performance starts to decrease.

\vspace{-1em}
\subsection{Robotic Grasping}
\label{sec:exp_robotic_grasping}
The final experiment evaluates the impact of active visuo-haptic object shape completion on grasp success rate. We used the same overall setup as in the real-world reconstruction experiment, but changed two of the objects. The ten objects we used are shown in \figref{fig:real_setup}. All of these objects, except the yellow mustard bottle, were completely new. We decided to benchmark \methodname{} against the Hull method because it reached the second-best reconstruction accuracy on simulated and real objects. 

For planning grasps on the reconstructed objects, we used the simulated annealing planner in \graspit{} that ran for 75000 steps. Out of the planned grasps, the first physically reachable grasp with the highest $\epsilon$-quality metric was executed on the robot. To study the effect of the number of touches on grasp success rate, we planned and executed a grasp after zero, three, and five touches. We repeated the reconstruction and grasping procedure three times for each combination of objects, method, and the number of touches, resulting in 180 grasps in total. The robot performed a grasp by first picking the object, then moving 10 cm upwards, and finally rotating the last joint $\pm 90^\circ$. The grasp was considered successful if the robot did not drop the object during this movement; otherwise, it was unsuccessful.  

\begin{figure}[tb]
    \centering
    \begin{subfigure}{0.195\textwidth}
        \centering
        \includegraphics[width=1\textwidth,trim={0 0 0cm 0cm},clip]{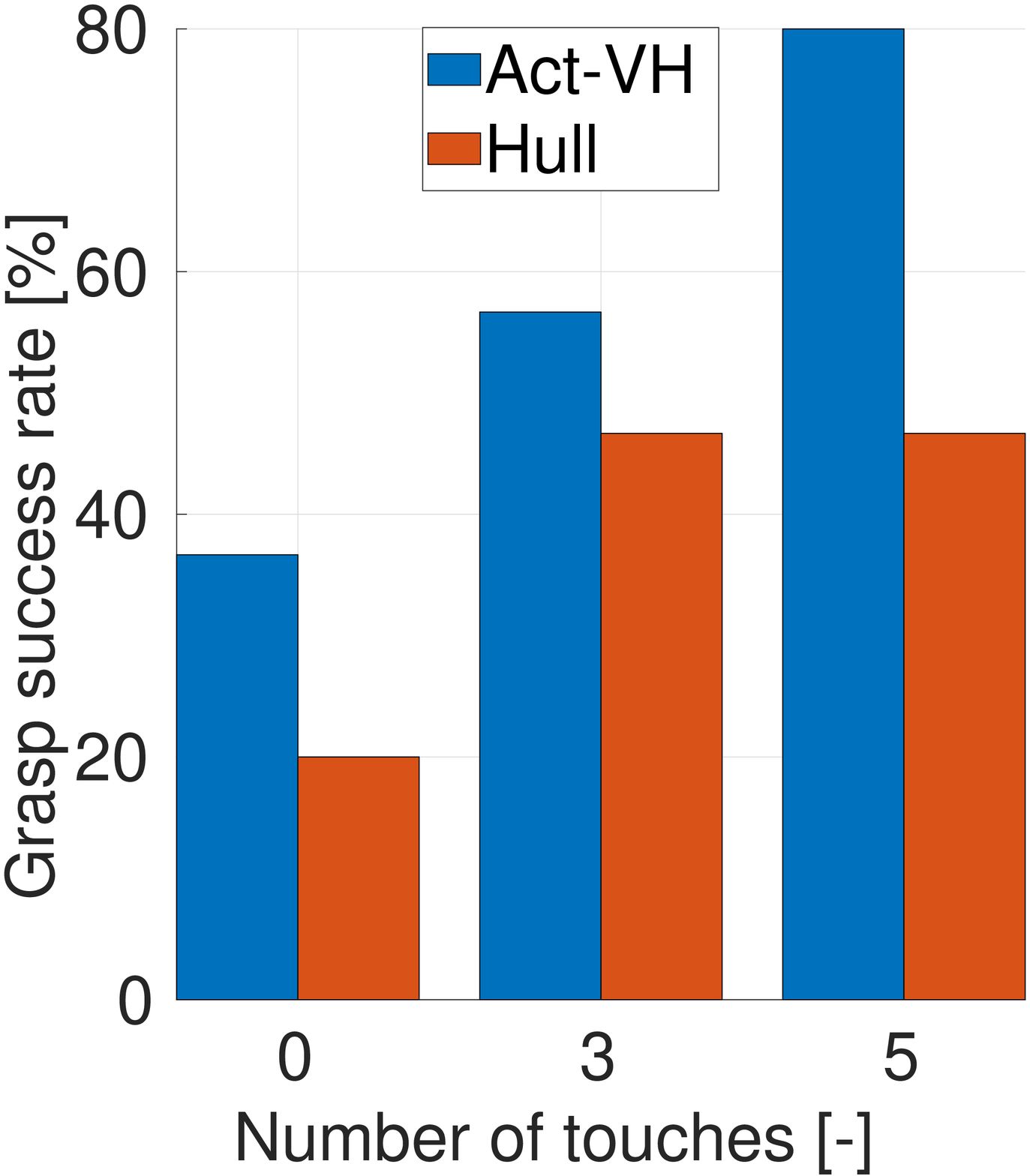}
        \caption{\label{subfigure:average_grasp_success_rate}}

    \end{subfigure}
    \begin{subfigure}{0.28\textwidth}
        \centering
        \includegraphics[width=0.9\textwidth,trim={0.5cm 0 1.0cm 0},clip]{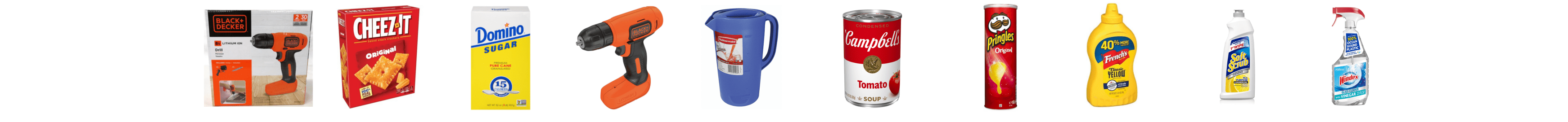}\\
        \includegraphics[width=1\textwidth,trim={0 0 0 0.5cm},clip]{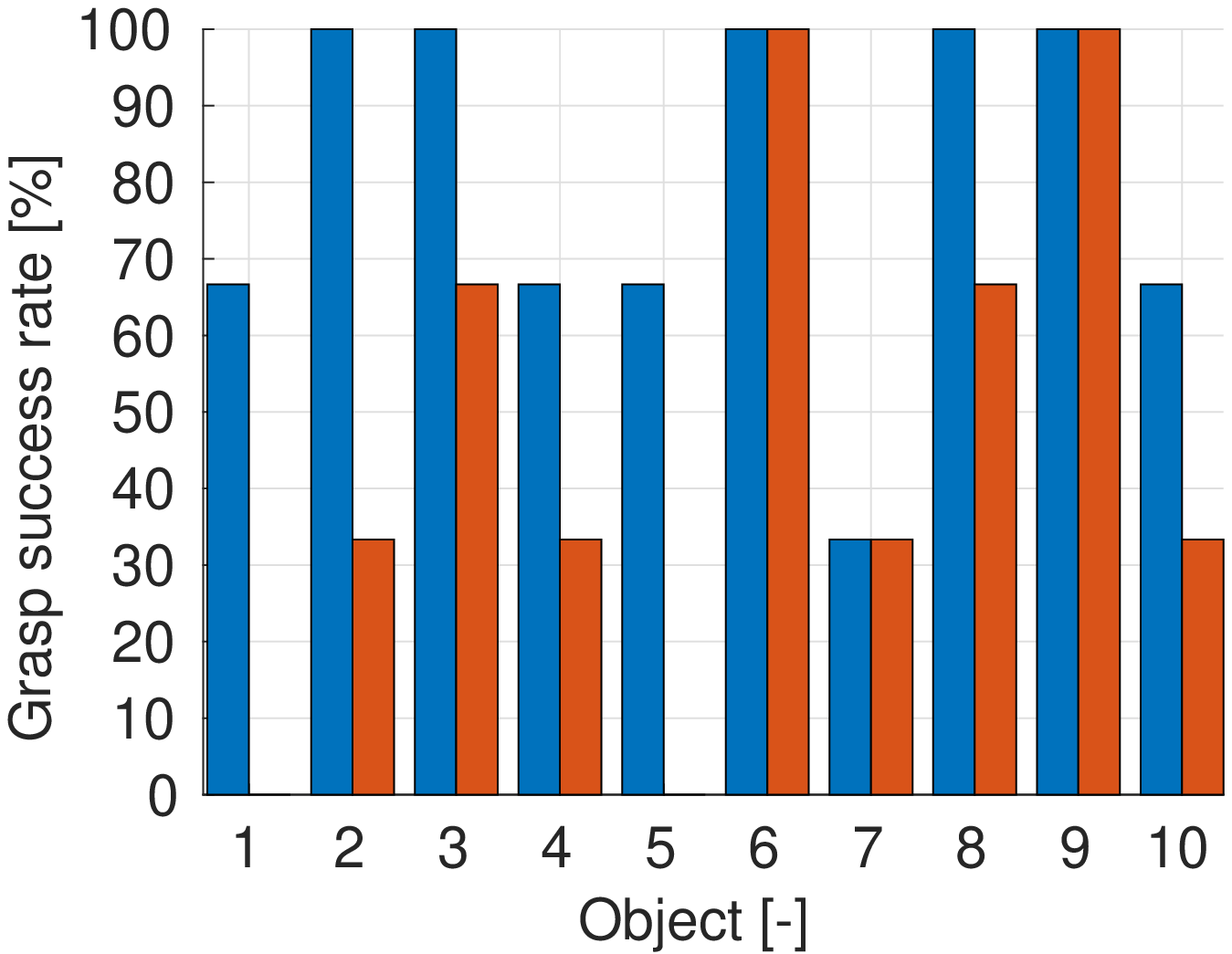}
        \caption{\label{subfigure:individual_average_grasp_success_rate}}

    \end{subfigure}
    \caption{Average grasp success rate of IGR (blue) and Hull (red).}
    \label{fig:grasping}
    \vspace{-2em}
\end{figure}

\figref{subfigure:average_grasp_success_rate} shows the average grasp success rates over varying number of touches. We can clearly see that \methodname{} is superior to Hull. For instance, after five touches, \methodname{} achieved an 80\% average grasp success rate while Hull only achieved 46.7\%. As expected, the grasp success rate with \methodname{} improves with the number of touches, from 38\% to 80\%. In comparison, the success rate for Hull was unchanged between 3 and 5 touches, indicating that the additional haptic data did not improve the convex hull reconstruction for grasp planning. This fact is highlighted in \figref{fig:meshes}, where the Hull reconstruction creates ``slopes'' from one set of points to another which the grasp planner seemed to favor but resulted in unsuccessful grasps. The same reconstruction artifacts were not visible for \methodname{}. Despite these differences, haptic exploration still increased the grasp success rate by more than 100\% between zero and five touches irrespective of the methods, highlighting the benefit of better reconstructions.

Finally, \figref{subfigure:individual_average_grasp_success_rate} shows the average grasp success rates of \methodname{} and Hull after five touches on the ten objects individually. The results indicate that \methodname{} performs better than or on par with Hull on all objects. Hull has particular problems with larger objects, such as objects 1, 2, 4, 5, and 10, which most probably stem from the ``slope'' artifacts mentioned earlier. \methodname{}, on the other hand, only performs poorly on object 7, where the two failed grasps were side-grasps for which the object slipped out of the gripper because of its short fingers. One possible reason why more side grasps were produced was because the reconstructed object was much thinner than in the real world (shown in the center row of \figref{fig:meshes}).    

%% file: Sections/conclusion.tex
\vspace{-1em}
\section{Conclusions and Future Work}

We presented \methodname{}, an active visuo-haptic shape completion method. The challenge in visuo-haptic shape completion is to decide the most informative touch location. To this end, \methodname{} uses a probabilistic shape completion network to assess where the reconstruction is most uncertain. This location is then used for haptic exploration. The reconstruction accuracy of \methodname{} compared to five baseline methods shows, both in simulation and real world, that \methodname{} produces the best reconstructions and that its reconstruction accuracy increases most with the number of haptic explorations. Furthermore, active visuo-haptic shape completion was also beneficial for robotic grasping where \methodname{} reached significantly higher grasp success rates than the Hull method. 

To assess the uncertainty of current shape reconstruction, we sampled latent codes during shape optimization and used the variance of the reconstructed shapes from this phase as a measure of uncertainty. Note that this method may not reflect the true uncertainty about the object shape given the available information. An alternative method---sampling and then optimizing multiple latent codes independently---did not yield better results and was computationally more expensive. However, this remains an empirical result and better theoretical grounding would be required.

Although \methodname{} achieved promising reconstruction accuracy and grasp success rates, there is still room for improvements. 
One improvement is to modify the loss function of IGR to also incorporate data points that we know are not on the surface. For instance, haptic exploration does not only indicate where the surface exists, but also where it does not exist. Another improvement is to also model the haptic location as uncertain. This would allow to select touch locations that are most robust to shape and robot uncertainties. Practically, performance would increase with a more sensitive contact detection method using e.g. a F/T sensor in the robot wrist or tactile sensors at the fingertip.

In summary, the work presented here shows that we can achieve accurate shape reconstructions with active visuo-haptic shape completion. This, in turn, enables the use of visuo-haptic exploration in perceptually uncertain environments such as cluttered scenes where objects occlude each other.